\documentclass{article}
\usepackage[utf8]{inputenc}
\usepackage{amsmath}
\usepackage{mathrsfs}
\usepackage{amsfonts}
\usepackage{mathtools}
\usepackage{tikz}
\usepackage{graphicx}

\title{Beyond Nyströmformer -  Approximation of  self -attention  by  Spectral Shifting }
\author{\textbf{Madhusudan Verma} \\ \\
Independent Researcher}

\date{vermamadhusudan2020@gmail.com}

\begin{document}
\maketitle
\begin{abstract}
Transformer is a powerful tool for many natural language tasks which is based on self-attention, a mechanism that encodes the dependence of other tokens on each specific token, but the computation of self-attention is a bottleneck due to its quadratic time complexity. There are various approaches to reduce the time complexity and approximation of matrix is one such. In Nyströmformer, the authors used Nyström based method for approximation of softmax. The Nyström method generates a fast approximation to any large-scale symmetric positive semidefinite (SPSD) matrix using only a few columns of the SPSD matrix. However, since the Nyström approximation is low-rank when the spectrum of the SPSD matrix decays slowly, the Nyström approximation is of low accuracy. Here an alternative method is proposed for approximation which has a much stronger error bound than the Nyström method. The time complexity of this same as Nyströmformer which is $O\left({n}\right)$.    
\end{abstract}
\maketitle

\section{Introduction and related work }
Transformer(Vaswani et al. 2017) has become the popular model for natural language processing including text classification, translation(Ott et al. 2018), or question answering system. Models that uses transformer have a huge number of parameters starting from 340 million in BERT-large to 175 billion in GPT-3. Due to this training and deploying such models are slow and require extensive distillation or compression to use for real-life applications.
 
The main bottleneck is self-attention which requires $O\left({n^2}\right)$ There were prior works done to reduce this complexity One method was to introduce sparsity into the attention layers by making each token to attend only a subset of tokens of an entire sequence. But this method suffers from a large performance drop with limited efficiency gain. Then Reformer was introduced which uses locally sensitive hashing was used to avoid costly computation they also proposed to use reversible layers to allow storing the only once instead of for each layer but its efficiency gain appears only after sequence length $\geq$ 2048.

\begin{center}
\begin{tabular}{ | p{5cm}| p{2cm}| } 
\hline
Model & Complexity  \\ 
\hline
Transformer & $O\left({n^2}\right)$ \\ 
\hline
Sparse Transformer & $O\left({n\sqrt{n}}\right)$  \\ 
\hline
Reformer & $O\left({n log(n)}\right)$  \\ 
\hline
Linformer & $O\left({n}\right)$ \\ 
\hline

Nyströmformer & $O\left({n}\right)$   \\
\hline

This method & $O\left({n}\right)$   \\

\hline
\end{tabular}
\end{center}
\section{Background}
\subsection{Self Attention}

Let there be n tokens with input dimension d ,$X \in R^{n\times d}$ is projected with matrices $W_Q$ $W_K$ $W_V$ to get Q,K,V knows as queries,keys and values respectively .Self attention is defined as follows(Vaswani et al. 2017)
$$S=softwmax\left(\frac{QK^T}{\sqrt{d_k}}\right)$$ where $d_k$ is the dimension of keys and $d_k=d_q$ where $d_q$ is the dimension of queries.
\subsection{Nyström Approximation}
Given 
 $S = \begin{bmatrix}
       A_s & B_s\\
       F_s & C_s \\
     \end{bmatrix}$
The Nyström approximation for matrix (Williams
and Seeger 2001) is given by ,
$\hat{S} = \begin{bmatrix}
       A_s \\
       F_s  \\
     \end{bmatrix}A_s^+\begin{bmatrix}
       A_s &&B_s
     \end{bmatrix}$
     for softmax $$\hat{S}=\left[softmax\left(\frac{QK^T}{\sqrt{d_k}}\right)\right]_{n\times m} A_s^+\left[softmax\left(\frac{QK^T}{\sqrt{d_k}}\right)\right]_{m\times n}$$
     where $[\quad]_{n\times m}$ refers to taking m columns from n × n matrix
and $\left[\quad\right]_{m\times n}$
refers to taking m rows from n × n matrix
Due to the fact that softmax is a row-wise softmax one needs to know all the columns,even though this method requires a subset of columns  and needs to consider landmark selection.
     
\subsection{Landmark selection}
Here we describe  Segment-means similar to the local average pooling previously used in the NLP literature (Shen et al.
2018a;Yunyang  et al. 2021). 
 For input queries Q 
 n queries are separated into m segments. As we can pad inputs to a length divisible to m, we assume n is divisible by m
for simplicity. Let $l = \frac{n}{m}$, landmark points for Q are computed in (1). Similarly, for input keys K,
landmarks are computed as shown in (1).
\begin{equation}\Tilde{q_j}=\displaystyle\sum_{i=(j-1)\times l+1}^{(j-1)\times l+m}\frac{q_i}{m}\quad \quad \quad \quad \quad  \Tilde{k_j}=\displaystyle\sum_{i=(j-1)\times l+1}^{(j-1)\times l+m}\frac{k_i}{m}\end{equation}
\subsection{Nyström approximation with landmark selection}

Instead of applying softmax and then selecting the columns here first the columns are selected using landmark and then the row wise softmax function is applied(Yunyang  et al. 2021) 

$$\hat{S}=softmax\left(\frac{Q\Tilde{K}^T}{\sqrt{d_k}}\right)softmax\left(\frac{\Tilde{Q}\Tilde{K}^T}{\sqrt{d_k}}\right)softmax\left(\frac{\Tilde{Q}K^T}{\sqrt{d_k}}\right) $$
\section{Spectral shifting method}

Let K $\in \boldsymbol{R}^{n\times n}$ and let P $\in \boldsymbol{R}^{n\times c}$ be the column selection matrix and $\Tilde{C}=\Tilde{K}{P}$  where 
$\Tilde{K}=K$ or K - $\delta I_n$
for some parameter $\delta \geq 0$
we approximate K by $\Tilde{C}U^{SS}\Tilde{C}^T           + \delta^{SS}I_n$
where ( Wang et.al 2016)
\begin{equation}
   \left(U^{SS},\delta^{SS}\right)=argmin_{U,\delta}\|K-\Tilde{C}U^{SS}\Tilde{C}^T - \delta^{SS}I_n\| 
\end{equation} 
This has closed form solution 
\begin{align*}
     \delta^{SS}& =\frac{1}{n-rank(\Tilde{C})}\left(tr(K)-tr(\Tilde{C}^\dagger K\Tilde{C})\right)\\
     U^{SS}&=\Tilde{C}^\dagger K(\Tilde{C}^\dagger)^T)  - \delta^{SS}(\Tilde{C}^T\Tilde{C})^\dagger
\end{align*}
Spectral shifting is more accurate than prototype model but it's complexity is $O(n^2c)$ This method also uses entire matrix so in case a matrix is a product of two matrix then one needs to the the product before selection of columns.We use an approach to use only a subset of columns for approximation

\section{Modified Spectral Shifting}
Let K $\in \boldsymbol{R}^{n\times n}$ and let P $\in \boldsymbol{R}^{n\times c}$ be the column selection matrix and $\Tilde{C}=\Tilde{K}{P}$  where 
$\Tilde{K}=K$ or $K - \delta I_n$
for some parameter $\delta \geq 0$
we approximate K by $\Tilde{C}U^{SS}\Tilde{C}^T           + \delta^{SS}I_n$
where 
\begin{equation}
   \left(U^{SS},\delta^{SS}\right)=argmin_{U,\delta}\|P^T(K-\Tilde{C}U^{SS}\Tilde{C}^T           - \delta^{SS}I_n)P\| 
\end{equation} 
This has closed form solution 
\begin{align*}
    \delta^{SS}& =\frac{1}{c-rank(A_s)}\left(tr(A_s)-tr(A_s^\dagger A_s^2)\right)\\
     U^{SS}&=A_s^\dagger   - \delta^{SS}\left({A_s^2}\right)^\dagger(\text{Since we assumed K= $K^T$ and $\Tilde{K}$=K})
\end{align*}
The time complexity of the above method is $O(c^3)$ .This method requires to consider only a subset of columns.\\ \\ \\ \\
\textbf{Lemma 1: } Let K be an $n\times n$ SPSD 
2
 matrix such that $\lambda_1(K)\geq\dots\geq\lambda_k(K)>\theta=\lambda_{k+1}=\dots=\lambda(K)>0.$By sampling $c=O(k)$ columns by the near-optimal $+$ adaptive algorithm we have that( Wang et.al 2016), $$\|K-\Tilde{K}^{SS}_c\|=0$$
\textbf{Theorem  1: }Modified Spectral Shifting is more accurate than Prototype model if K and c follows conditions mentioned in Lemma 1 \\
\textbf{Proof: }
\begin{align*}
   \|P^T(K-\Tilde{C}U^{SS}\Tilde{C}^T           - \delta^{SS}I_n)P\|
   \leq \|P^T\|\|(K-\Tilde{C}U^{SS}\Tilde{C}^T           - \delta^{SS}I_n)\|\|P\|
    = 0 \\ \text{ (by using lemma 1)} \\
 \text{But by definition of norm , } 
 \|P^T(K-\Tilde{C}U^{SS}\Tilde{C}^T  - \delta^{SS}I_n)P\| \geq 0  \text{ Therefore ,} \\
 \|P^T(K-\Tilde{C}U^{SS}\Tilde{C}^T - \delta^{SS}I_n)P\|= 0 \leq \|K-K_c^\text{nyst}\| \\ \text{(Since norm is always greater than zero by definition)}
\end{align*}

\section{ Modified Spectral Shifting method using landmarks  }
We cannot apply the modified Spectral shifting method directly on the $softmax\left(\frac{QK^T}{\sqrt{d_k}}\right) $ matrix  even thought it uses only a subset of columns because of the row-wise softmax function depicted in figure 1.
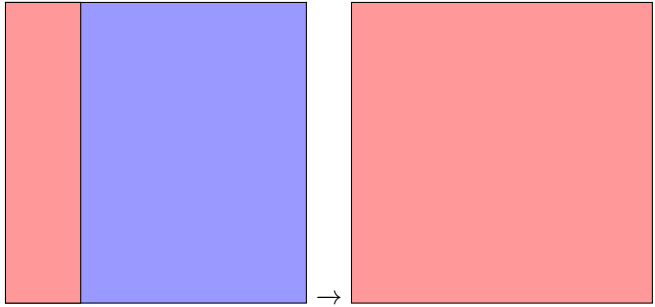
\begin{figure}
     
     \begin{tikzpicture}
\filldraw[fill=blue!40!white, draw=black] (0,0) rectangle (4,4);
\filldraw[fill=red!40!white, draw=black] (0,0) rectangle (1,4) ;
\end{tikzpicture} $\xrightarrow{}$
\begin{tikzpicture}
\filldraw[fill=red!40!white, draw=black] (0,0) rectangle (4,4) ;
\end{tikzpicture}
     \caption{The red rectangle represents  the columns used for modified Spectral Shifting method.The blue rectangle represents the $\frac{QK^T}{\sqrt{d_k}}$.The key challenge for applying modified spectral shifting directly is computing the submatrix requires all the entries $n\times n$ before applying softmax .Therefore it has the same complexity of $O(n^2)$ .}
     \label{fig: modified}
 \end{figure}

So instead of applying row-wise softmax function first and then selecting columns we first select columns using landmark and the apply row-wise softmax function on that matrices formed after landmark selection

Let $\mathcal{L}$ denote the row-wise softmax function, $\Tilde{Q}$ and $\Tilde{K}$ are matrices formed by landmark selection from Q and K respectively ,then the approximation can be written as
\begin{equation}
    \mathcal{L}\left(\frac{Q\Tilde{K}^T}{\sqrt{d_k}}\right) \mathcal{L}\left(\frac{\Tilde{Q}\Tilde{K}^T}{\sqrt{d_k}}\right)^\dagger\left(I_c-\delta^{SS} \mathcal{L}\left(\frac{\Tilde{Q}\Tilde{K}^T}{\sqrt{d_k}}\right)\right) \mathcal{L}\left(\frac{\Tilde{Q}K^T}{\sqrt{d_k}}\right)
\end{equation}
\textbf{Proof: }
Let $A_s=softmax\left(\frac{\Tilde{Q}\Tilde{K}^T}{\sqrt{d_k}}\right)$ where $A_s=U_{c\times c}\Gamma_{c\times c}V_{c\times c}$ \\ for a given query $q_i$ and key $k_j$
let $A_{\Tilde{K}}(q_i)=softmax\left(\frac{q_i\Tilde{K}^T}{\sqrt{d_k}}\right)$;   $A_{\Tilde{Q}}(k_j)=softmax\left(\frac{\Tilde{Q}k_j^T}{\sqrt{d_k}}\right)$ where $A_{\Tilde{K}}(q_i)\in R^{1\times c}$ and $A_{\Tilde{K}}(k_j)\in R^{c\times 1}$  we can construct two vectors 
\begin{align*}
   \textbf{x}_{\Tilde{K}}(q_i)&=\left(I_c^\frac{1}{2}-\Gamma_{c\times c}^{-\frac{1}{2}}V^T\right)V^TA_{\Tilde{K}}(q_i)^T \\
   \textbf{x}_{\Tilde{Q}}(k_j)&=\left(I_c^\frac{1}{2}-\Gamma_{c\times c}^{-\frac{1}{2}}U^T\right)U^TA_{\Tilde{Q}}(k_j)^T
\end{align*}
so the entries of $\Tilde{S}$ can be calculated using landmark matrices $\Tilde{K}$ and $\Tilde{Q}$ and is given by
$$ \Tilde{S}_{ij}=\textbf{x}_{\Tilde{K}}(q_i)^T\textbf{x}_{\Tilde{Q}}(k_j) ,  \forall i=1,\dots,n,j=1,\dots,n $$
In order to derive the method we assume that $A_s$ to be non-singular to define the above two vectors meaningful then we can relax the assumption by replacing with pseudo inverse
When $A_s$ is non-singular
\begin{align}
    \Tilde{S}_{ij}&=\textbf{x}_{\Tilde{K}}(q_i)^T\textbf{x}_{\Tilde{Q}}(k_j)\\
    &=A_{\Tilde{K}}(q_i)V_{c\times c}\Gamma_{c\times c}^{-1}U_{c\times c}^T\left(I_c-\delta^{SS}V_{c\times c}\Gamma_{c\times c}^-1U_{c\times c}^T\right)A_{\Tilde{Q}}(k_j)\\
    &=A_{\Tilde{K}}(q_i)A_s^-1\left(I_c-\delta^{SS}A^{-1}_s \right)A_{\Tilde{Q}}(k_j)
\end{align}
but if $A_s$ is non singular we can write
\begin{align}
    \Tilde{S}_{ij}=A_{\Tilde{K}}(q_i)A_s^\dagger\left(I_c-\delta^{SS}A^{\dagger}_s \right)A_{\Tilde{Q}}(k_j)
\end{align}
So,
\begin{equation}
  \Tilde{S}_{ij}=\mathcal{L}\left(\frac{q_i\Tilde{K}^T}{\sqrt{d_k}}\right) \mathcal{L}\left(\frac{\Tilde{Q}\Tilde{K}^T}{\sqrt{d_k}}\right)^\dagger\left(I_c-\delta^{SS} \mathcal{L}\left(\frac{\Tilde{Q}\Tilde{K}^T}{\sqrt{d_k}}\right)\right) \mathcal{L}\left(\frac{\Tilde{Q}k_j^T}{\sqrt{d_k}}\right)  
\end{equation}
for $i$,$j$ = {1,\dots,n}, $S$ is approximated as 
\begin{equation}
    \Tilde{S}=\mathcal{L}\left(\frac{Q\Tilde{K}^T}{\sqrt{d_k}}\right) \mathcal{L}\left(\frac{\Tilde{Q}\Tilde{K}^T}{\sqrt{d_k}}\right)^\dagger\left(I_c-\delta^{SS} \mathcal{L}\left(\frac{\Tilde{Q}\Tilde{K}^T}{\sqrt{d_k}}\right)\right) \mathcal{L}\left(\frac{\Tilde{Q}K^T}{\sqrt{d_k}}\right) 
\end{equation}
\section{Experiments}
Here we show that  approximation matrix  method is not   low rank compared to previous  method  and therefore is better approximation .
\begin{figure}
    \centering
    \includegraphics{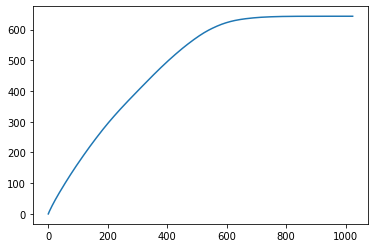}
    
    \includegraphics{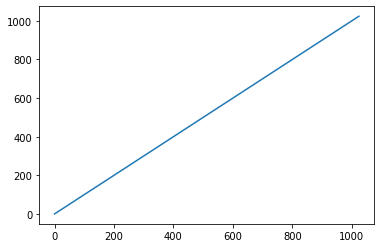}
    \caption{Spectrum analysis  of sefl attention matix(top) and  approximate matrix(bottom), y axis represents cumulative eigen value and x axis represents   eigen value index.We can see that bottom image has no long tail so it is not a low rank matrix }
    \label{fig:my_label}
\end{figure}
\section{Error bound}
Let $Z^*$ be the approximate pseudo inverse  of $A_s$by using iterative method given  by
\begin{equation}
    Z_{j+1}=\frac{1}{4}
Z_j(13I - A_sZ_j (15I - A_sZ_j )(7I - A_sZ_j)
\end{equation}
     with initial approximation $Z_0$  satisfying  $\|A_sA_s^\dagger
 -
A_sZ_0\| < 1$ 
\text{ then, }
\begin{equation}
    E\leq 1+\|A_s^\dagger\|_\infty(1+\delta^{SS}\|A_s^\dagger\|_\infty)(1-\|A_s^\dagger-Z^*\|_\infty)
\end{equation}
\textbf{Proof: }

\begin{align*}
     E&=\|\mathcal{L}\left(\frac{QK^T}{\sqrt{d_k}}\right)-\mathcal{L}\left(\frac{Q\Tilde{K}^T}{\sqrt{d_k}}\right)Z^*\left(I_c-\delta^{SS}\mathcal{L}\left(\frac{\Tilde{Q}K^T}{\sqrt{d_k}}\right)\right)\mathcal{L}\left(\frac{\Tilde{Q}K^T} {d_k}\right)\|_\infty\\
     &=\|\mathcal{L}\left(\frac{QK^T}{\sqrt{d_k}}\right)-\mathcal{L}\left(\frac{Q\Tilde{K}^T}{\sqrt{d_k}}\right)\mathcal{L}\left(\frac{\Tilde{Q}\Tilde{K}}{d_k}\right)^\dagger\left(I_c-\delta^{SS}\mathcal{L}\left(\frac{\Tilde{Q}K^T}{\sqrt{d_k}}\right)\right)\mathcal{L}\left(\frac{\Tilde{Q}K^T} {d_k}\right)\\
     &+\mathcal{L}\left(\frac{Q\Tilde{K}^T}{\sqrt{d_k}}\right)\mathcal{L}\left(\frac{\Tilde{Q}\Tilde{K}}{d_k}\right)^\dagger\left(I_c-\delta^{SS}\mathcal{L}\left(\frac{\Tilde{Q}K^T}{\sqrt{d_k}}\right)\right)\mathcal{L}\left(\frac{\Tilde{Q}K^T} {d_k}\right) \\
     &-\mathcal{L}\left(\frac{Q\Tilde{K}^T}{\sqrt{d_k}}\right)Z^*\left(I_c-\delta^{SS}\mathcal{L}\left(\frac{\Tilde{Q}K^T}{\sqrt{d_k}}\right)\right)\mathcal{L}\left(\frac{\Tilde{Q}K^T} {d_k}\right)\|_\infty\\
     &\overset{(a)}{\leq}
     \|\mathcal{L}\left(\frac{QK^T}{\sqrt{d_k}}\right)\|_\infty\\ &+\|\mathcal{L}\left(\frac{Q\Tilde{K}^T}{\sqrt{d_k}}\right)\mathcal{L}\left(\frac{\Tilde{Q}\Tilde{K}}{d_k}\right)\left(I_c-\delta^{SS}\mathcal{L}\left(\frac{\Tilde{Q}K^T}{\sqrt{d_k}}\right)\right)\mathcal{L}\left(\frac{\Tilde{Q}K^T} {d_k}\right)\|_\infty\\
     &+\|\mathcal{L}\left(\frac{Q\Tilde{K}^T}{\sqrt{d_k}}\right)\left(I_c-\delta^{SS}\mathcal{L}\left(\frac{\Tilde{Q}K^T}{\sqrt{d_k}}\right)\right)\mathcal{L}\left(\frac{\Tilde{Q}K^T} {d_k}\right)\|_\infty\|\mathcal{L}\left(\frac{\Tilde{Q}\Tilde{K}}{d_k}\right)^\dagger-Z^*\|_\infty\\
     &\overset{(b)}{\leq}
    \|\mathcal{L}\left(\frac{QK^T}{\sqrt{d_k}}\right)\|_\infty + \|\mathcal{L}\left(\frac{Q\Tilde{K}^T}{\sqrt{d_k}}\right)\|_\infty
    \| \mathcal{L}\left(\frac{\Tilde{Q}\Tilde{K}^T}{\sqrt{d_k}}\right)^\dagger\|_\infty\|I_c-\delta^{SS} \mathcal{L}\left(\frac{\Tilde{Q}\Tilde{K}^T}{\sqrt{d_k}}\right)^\dagger\|_\infty \\
    &\left(\left\|\mathcal{L}
    \left(\frac{\Tilde{Q}K^T}{\sqrt{d_k}}\right)\right\|_\infty -\left\|\mathcal{L}\left(\frac{\Tilde{Q}\Tilde{K}^T}{\sqrt{d_k}}\right)^\dagger-Z^*\right\|_\infty\right)\\
    &\overset{(c)}{\leq}
    1+\|A_s^\dagger\|_\infty(1+\delta^{SS}\|A_s^\dagger\|_\infty)(1-\|A_s^\dagger-Z^*\|_\infty)
\end{align*}

 Step (a) uses triangle inequality and submultiplicative property of norm of matrices.Step (b) uses submultiplicative property.
 Step (c) uses the fact that $\|\mathcal{L}(A)\|_\infty$ for any matrix A.
 
 \section{Time complexity Analysis }
 We now provide a complexity analysis of this  approximation method which needs to account ¨
for landmark selection, pseudoinverse calculation, and the
matrix multiplications. Landmark selection using Segmentmeans takes $O(n)$. Iterative approximation of the pseudoinverse takes $O(c^3)$ in the worst case.Here we first compute $\mathcal{L}\left(\frac{Q\Tilde{K}^T}{\sqrt{q_k}}\right) \times Z^* $ then calculates $\mathcal{L}\left(\frac{\Tilde{Q}K^T}{\sqrt{q_k}}\right) \times V $ and finally $\mathcal{L}\left(\frac{Q\Tilde{K}^T}{\sqrt{q_k}}\right) \times Z^* \times \mathcal{L}\left(\frac{\Tilde{Q}K^T}{\sqrt{q_k}}\right) \times V$.This costs $O(nc^2 +cnd_k +c^3 +ncd_k)$. The overall time
complexity is thus $O(c^3 + nc^2 +cnd_k +c^3 +ncd_k)$.Thus this scales linearly with respect to input sequence length n.
\section{Conclusion}
In this paper we have seen how modified Sprectral Shifting method can be used to approximate self attention in linear time.This can be used for calculating self attention for long sequences in Transformer as well as vision tasks.Current transformer models takes huge computation time for training and inference which limits its practical applications .Even if one train using computational resources then also deployment is issue.This method can reduce training and inference time.

\end{document}